\def\year{2016}\relax
\documentclass[letterpaper]{article}
\usepackage{aaai16}
\usepackage{times}
\usepackage{helvet}
\usepackage{courier}
\usepackage{graphicx}
\usepackage{wrapfig}
\usepackage{url}
\usepackage{amsmath}
\usepackage[font={small}]{caption}
\usepackage{color}
\usepackage{amsmath,amssymb}
\usepackage[font={small}]{caption}
\usepackage{enumitem}
\usepackage{bm}

\newcommand\numberthis{\addtocounter{equation}{1}\tag{\theequation}}

\begin{document}

\title{Modeling Language Vagueness in Privacy Policies using Deep Neural Networks}

\author{
$^1$Fei Liu, $^2$Nicole Lee Fella, $^1$Kexin Liao\\
$^1$University of Central Florida, 4000 Central Florida Blvd., Orlando, Florida 32816\\
$^2$Manhattan College, 4513 Manhattan College Parkway, Riverdale, NY 10471\\
{\tt feiliu@cs.ucf.edu, nfella01@manhattan.edu, ericaryo@knights.ucf.edu}
}
\maketitle

\begin{abstract}

Website privacy policies are too long to read and difficult to understand.
The over-sophisticated language makes privacy notices to be less effective than they should be.
People become even less willing to share their personal information when they perceive the privacy policy as vague.
This paper focuses on decoding vagueness from a natural language processing perspective.
While thoroughly identifying the vague terms and their linguistic scope remains an elusive challenge,
in this work we seek to learn vector representations of words in privacy policies using deep neural networks.
The vector representations are fed to an interactive visualization tool (\textsf{LSTMVis}) to test on their ability to discover syntactically and semantically related vague terms.
The approach holds promise for modeling and understanding language vagueness.

\end{abstract}

\section{Introduction}

Website privacy policies represent a legally binding agreement between the users and website operators.
They are verbose, too long to read, and difficult to understand.
Albeit the paramount importance, people tend to ignore these privacy notices unless a serious concern is raised, e.g., by the media.
Research studies have explored various means to improve the effectiveness of privacy notice and choice.
Cranor et al.~\shortcite{Cranor:2002,Cranor:2006} introduce a standard machine-readable format for website privacy policies using the Platform for Privacy Preferences (P3P).
The Usable Privacy Policy Project\footnote{\url{https://www.usableprivacy.org}} aims to extract key privacy policy features for presentation to end-users in a structured and easily understandable format~\cite{Sadeh:2013}. 
These approaches allow the users to quickly jump to the text passages that are related to certain key privacy practices.
It alleviates the ``too long to read'' challenge brought by document length.
On the other hand, studies that tackle the ``difficult to understand'' challenge have been largely absent from this space, partly because of the complexity and richness of natural language.

One might wonder why privacy notices need to adopt such sophisticated language in the first place.
Bhatia et al.~\shortcite{Bhatia:2016} suggest two causes in their recent work.
First, the policies need to be \textit{comprehensive}, covering all possible cases such as the physical places (e.g., stores, offices) and web/mobile platforms. 
Second, the policy statements must be \textit{accurate}, which means they are true to all data practices and systems.
Clearly it will be difficult for the legal counsel to anticipate all the future needs, 
naturally they resort to generalization and sophistication to frame the statements, introducing vagueness to the text.
An example statement is: 
%``\textit{However, it is possible that some parts of the Web Sites will not operate correctly if you disable cookies and you may not be able to take advantage of some of this Web Sites' features.}''
``\textit{The email address is used for sending account notifications and other system-related information as needed.}''

\begin{table}[t]
\setlength{\tabcolsep}{7pt}
\renewcommand{\arraystretch}{1.2}
\begin{tabular}{|l|}
\hline
\textbf{Condition (9)}: depending, necessary, appropriate, \\
inappropriate, as needed, as applicable, otherwise \\
reasonably, sometimes, from time to time\\
\hline
\textbf{Generalization (12)}: generally, mostly, widely, general,\\
commonly, usually, normally, typically, largely, often,\\
primarily, among other things\\
\hline
\textbf{Modality (8)}: may, might, can, could, would, likely,\\
possible, possibly\\
\hline
\textbf{Numeric Quantifier (11)}: anyone, certain, everyone, \\
numerous, some, most, few, much, many, various,\\
including but not limited to\\
\hline
\end{tabular}
\caption{Table adopted from~\cite{Reidenberg:2016}. It includes a total of 40 vague terms that are manually identified by experts from 15 privacy policies. The terms are divided into four categories. }
\label{tab:vague_terms}
\vspace{-0.1in}
\end{table}

Vagueness is a linguistic phenomenon that is not yet fully studied in the natural language processing (NLP) community.
A concept is considered vague if it lacks clarity or corresponds to borderline cases (e.g., tall, short). 
Even terms like ``disability'' raise questions such as ``how much loss of vision is required before one is legally blind?''\footnote{https://en.wikipedia.org/wiki/Vagueness}
Farkas et al.~\shortcite{Farkas:2010} introduce a shared task on detecting uncertainty cues (i.e., hedges and weasels) from biological articles and Wikipedia pages.
Reidenberg et al.~\shortcite{Reidenberg:2016} manually analyze a set of 15 privacy policies and identify 40 vague terms (Table~\ref{tab:vague_terms}) which we also adopt in this study.
Note that there appears to be a dilemma:
if a collection of vague terms can be pre-specified, classifying a piece of text as vague or not seems trivial;
on the other hand, given the richness of natural language, creating such a comprehensive list of vague terms can be highly challenging, if possible at all.

%There is no consistent guideline.
%Vagueness is prevalent in daily language but undesirable in the legal documents.

The main contribution of this work lies in learning vector representations for words in privacy policies using deep neural networks.
There is one vector representation for each word token in the privacy policies.
The vector representations are iteratively learnt so that they would perform well in two tasks: predicting the next word given its context (e.g., ``\textit{we do not request any \_\_\_\_}'' $\rightarrow$ ``information'') and predicting whether or not a word is in a list of pre-specified vague terms (e.g., ``\textit{may}'' $\rightarrow$ ``Vague (V)'', ``\textit{email}'' $\rightarrow$ ``Not Vague (N)''.
The 40 vague terms in Table~\ref{tab:vague_terms} are used in this study.
We hypothesize that certain dimensions of the vector representation encode the semantic meaning of the word, including vagueness.
These vector representations are further fed to an interactive visualization tool (\textsf{LSTMVis}) to test on their ability to discover semantically related terms. 
The approach holds promise for modeling vagueness of words within context.
The visualization tool allows the privacy researchers to perform knowledge discovery on the website privacy policies.\footnote{We plan to release the code and data models upon acceptance of the paper.}

\section{Related Work}
\label{sec:related}

We discuss related work along three dimensions: law, privacy policy, and natural language.

In the American Constitution, there is a ``void for vagueness'' doctrine.
It states that the law should be clearly specified so that the average citizen would understand.
If a rule is vague, it is unenforceable. 
Researchers in the law community have thus exploited the vagueness of legal language and interpretation of boundary decisions.
In his seminal work, Waldron~\shortcite{Waldron:1994} distinguishes ambiguity, contestability, vagueness, and introduce a general term ``indeterminacy'' to cover the three cases.
Post~\shortcite{Post:1994} argues that the legal rules cannot be simply rewritten to be more precise, since they are not in isolation but reflect the forms of social order. 
Jonsson~\shortcite{Jonsson:2009} suggests that vagueness in law does not call for specific interpretation of the law itself, but only for an application of the law on case-by-case basis.
Studies in~\cite{Hernacki:2012} suggest that the decades-old antihacking statue Computer Fraud and Abuse Act (CFAA) is in need of a face-lift. Phrases such as ``involve'' and ``other similar information'' are not providing enough clarity.
Liebwald~\shortcite{Liebwald:2013} concerns that the vagueness in combination with the elasticity of legal interpretation may affect the binding force of law. The paper introduces a theory called Hyperbola of Meaning.
%More recent studies start to explore the language perspective of vagueness.
Raffman~\shortcite{Raffman:2015} provides a characterization of linguistic vagueness. Vague words possess unclear boundaries, but are distinguished from ambiguity, underspecificity, and several forms of indeterminacy. 
Low et al.~\shortcite{Low:2015} illustrate the application of the vagueness doctrine to four Supreme Court vagueness cases. They point out that when determining vagueness of statutes, it is important to take the intersection between state and federal law into account. 
Hunt~\shortcite{Hunt:2015} studies ``epistemicism'', which states that vague statements are either true or false even though it is impossible to know which. The author suggests that vagueness should be explained within the theory of legal interpretation.
%The law is expected to clearly distinguish between acceptable and forbidden behaviors so as both to guide the actions of citizens and to restrict the discretion of government officials.~\cite{}

Vagueness has been studied within the scope of website privacy policies.
In particular, Reidenberg et al.~\shortcite{Reidenberg:2016} and Bhatia et al.~\shortcite{Bhatia:2016} introduce a theory of vagueness for privacy policy statements.
The theory indicates how vague modifiers can be composed to increase or decrease the overall vagueness.
They show that the increases in vagueness often decrease users' willingness to share personal information.
Our work is different from these studies in that we do not attempt to generate a vagueness score for a given piece of text. 
Instead, we seek to exploit deep neural networks to learn word representations that encode semantic meanings and vagueness.

There are other studies that focus on improving the effectiveness of privacy policies.
Vail et al.~\shortcite{Vail:2008} compare various ways to present privacy policy information to online users. Their findings suggest that users perceive paragraph-form policies to be more secure than others, however the user comprehension of such paragraph-form policies is poor.
Kelley et al.~\shortcite{Kelley:2010} develop a nutrition label approach for representing the key practices of privacy policies. 
They show that a standardized table format is effective in assisting users with their information needs.
Micheti et al.~\shortcite{Micheti:2010} aims to identify guidelines for privacy plicies that children and teen can understand and accurately interpret.
Phrases which  cause misunderstanding and vagueness include ``may,'' ``except,'' and ``aside from.'' Many users, especially young people, are aware that privacy policies are vague due to strategic reasons of service providers.
The empirical study in~\cite{Lammel:2013} discusses Platform for Privacy Preferences (P3P). As a platform, P3P is used and interpreted by users to help automate decision making. While being one of the only widely used languages for privacy policies, P3P still has downfalls. More rigorous specifications in language and enforcement of correct use are necessary.
In more recent studies, Reidenberg et al.~\shortcite{Reidenberg:2015,Reidenberg:2015:Journal} study the effectiveness of privacy notice and choice framework and suggest that people do not agree with each other when interpreting privacy polices.
Wilson et al.~\shortcite{Wilson:2016,Wilson:2016:ACL} explore crowdsourcing for annotating privacy policies.
They introduce a corpus of 115 privacy policies with manual annotations of fine-grained data practices. 

Comprehensive studies have been missing for understanding vagueness in the natural language processing community.
Farkas et al.~\shortcite{Farkas:2010} focus on the detection of uncertainty cues and their linguistic scope in natural language texts.
The motivation behind this task is to distinguish factual and uncertain information in text, which is of essential importance to information extraction.
Much of the techniques involve sentence-level classifications using SVM, CRF, and maximum entropy.
However, it remains to be seen if a word- or sentence-level classification formulation is well suited for this task.
In~\cite{Alexopoulos:2014}, vagueness is considered to be a linguistic phenomenon. It arises with a lack of clear boundaries and conditions. These boundaries usually do not allow concrete distinction. Classifying text as vague or not vague can be subjective, making it important to look at agreement between interpretations and annotations. Using a naive Bayes classifier, the study shows that vague and not vague senses can be separated.

\section{Data}
\label{sec:data}

Our dataset consists of 1,010 website privacy policies.
These privacy policies are gathered using Amazon Mechanical Turk (\url{mturk.com}) from the most frequently visited websites across 15 categories, ranging from Arts, Business, Computers to Science, Shopping, and Sports.
The privacy policy documents have been converted to the XML format and are available for download publicly.\footnote{https://usableprivacy.org/data}
Liu et al.~\shortcite{Liu:2014} and Ramanath et al.~\shortcite{Ramanath:2014} 
perform studies using this dataset to align policy segments based on the issues they discuss.
For example, text segments that discuss the usage of cookies (i.e., small data files transferred by the website to the user's computer) should be grouped together.
They experiment with unsupervised hidden Markov models and demonstrate that the approaches are more effective than clustering.

In this study, we seek to learn a vector representation for each word token (i.e., occurrence of the word) in the dataset.
Each privacy policy document is split into a set of sentences;
each sentence is further split into a set of word tokens.
All word tokens are lowercased.
We remove sentences that contain 3 word tokens or less, since they are too short and often noisy (e.g., ``Back to Top'').
A special token $\langle/\mathsf{s}\rangle$ is used as the end-of-sentence symbol.
A word token is deemed \textit{vague} if it is included in a pre-specified list of vague terms (see Table~\ref{tab:vague_terms}).
Note that we consider word tokens such as ``\textit{among other things}'' as vague, but an individual word (e.g., ``\textit{among}'') is not vague when placed in other context (e.g., ``\textit{among consumers}'').

Statistics of the dataset are illustrated in Table~\ref{tab:dataset}.

%In total, we are able to gather a collection of 107,076 sentences.
%The average length of the sentences are 24 words.
%We set the maximum length of our sequence model to be 50 words. This correspond to 93.9\% of the total sentences.
%When using maxlen of 30 words, this corresponds to 73.3\% of total sentences.

\begin{table}[h]
\setlength{\tabcolsep}{7pt}
\renewcommand{\arraystretch}{1.2}
\centering
\begin{tabular}{|l|r|}
\hline
total \# of web privacy policies & 1,010 \\
total \# of sentences & 107,076 \\
total \# of word tokens & 2,534,094\\
total \# and \% of vague tokens & 59,026 (2.3\%)\\
total \# and \% of sentences that & \\
contain at least one vague token & 41,033 (38.3\%)\\
\hline
\end{tabular}
\caption{Statistics of the dataset.}
\label{tab:dataset}
\vspace{-0.1in}
\end{table}

%total vague terms: 59026 (2.3\%)
%total vague sentences: 41033 (38.3\%)
%total terms: 2534094

\section{Modeling Language and Vagueness}
\label{sec:model}

\begin{figure}[t]
\centering
\includegraphics[width=2.5in]{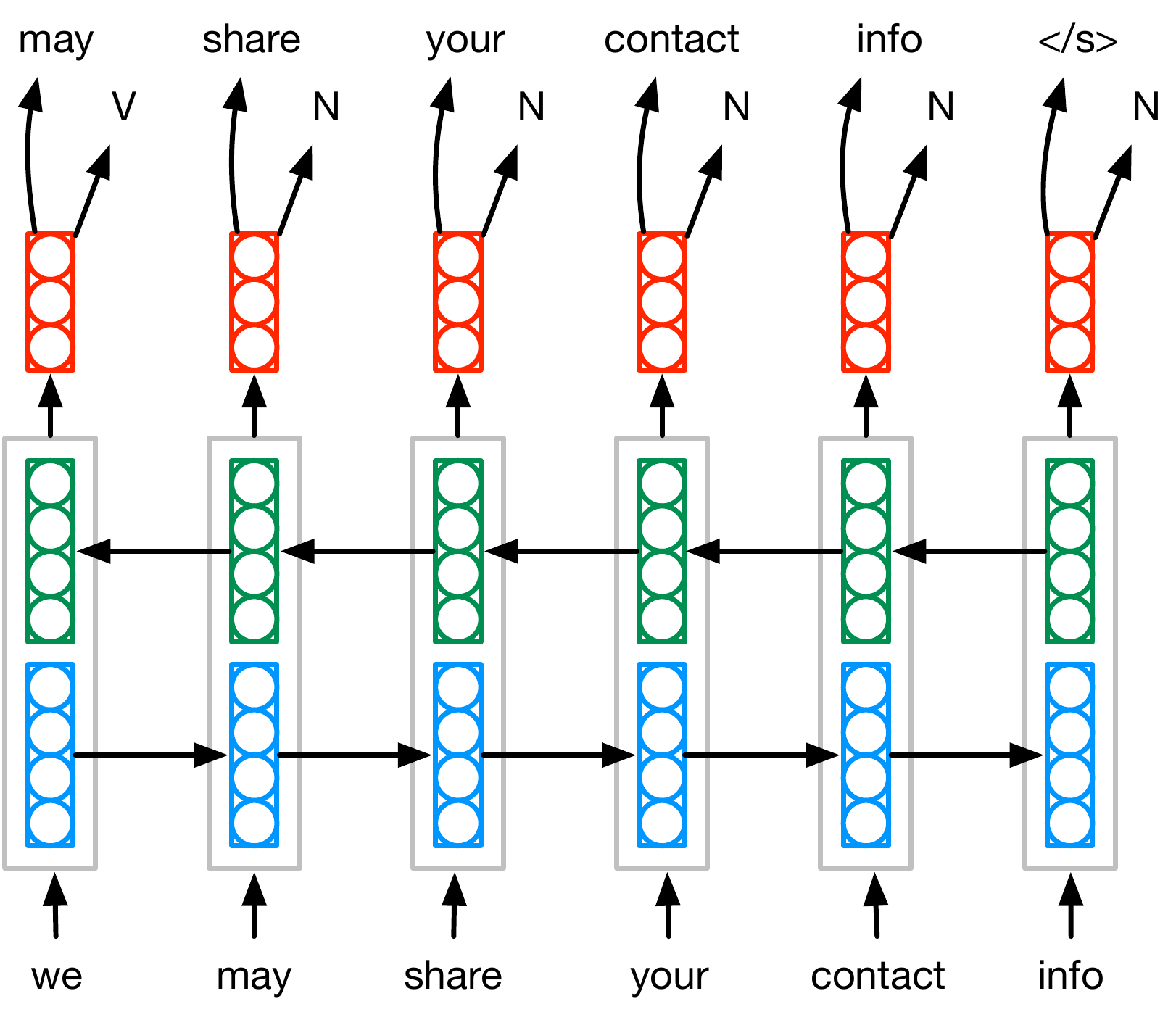}
\caption{A deep neural network for modeling language and vagueness in website privacy policies. ``Vague'' $\rightarrow$ V, ``Not Vague'' $\rightarrow$ N.}
\label{fig:model}
\vspace{-0.1in}
\end{figure}

So far we have demonstrated the needs for understanding language vagueness and described our dataset, we proceed by introducing a deep neural network for learning vector representations for words in privacy policies (see Figure~\ref{fig:model} for illustration).
Traditional approaches to building feature representation have been largely based on manual feature extraction~\cite{Farkas:2010}.
The idea behind the deep neural network is that it learns to automatically construct a feature representation for each word, in the form of a dense continuous vector ($\bm{g} \in \mathbb{R}^d$).
The feature representation is optimized so that it could perform well in two tasks: 1) predicting the next word given previous words in the sentence, and 2) predicting if the current word is vague or not given the context.
This corresponds to a multi-task learning setting.

Deep neural networks have seen considerable success in a range of natural language processing tasks.
Our work is inspired by recent advances on learning word embeddings~\cite{Mikolov:2013,Tang:2014} and sequence-to-sequence models~\cite{Sutskever:2014,Luong:2016}.

Concretely, let $\bm{x} = \{x_1, x_2, \cdots, x_\textsf{N}\}$ be an input sentence consisting of $\textsf{N}$ word tokens. 
The word tokens come from a vocabulary $\mathcal{V}$ of size $|\mathcal{V}| = \textsf{V}$. 
Each word is replaced by a pre-trained word embedding ($\bm{x}_i \in \mathbb{R}^\textsf{D}$) before it is fed to the neural network. 
With a slight abuse of notation, we use $x_i$ to represent the word token and $\bm{x}_i$ (bold-face) to represent its embedding.
We use the 300-dimension ($\textsf{D}=300$) \textsf{word2vec} embeddings pre-trained on Google News dataset with about 100 billion words\footnote{https://code.google.com/archive/p/word2vec/}.
A vocabulary of 5,000 words is employed in this study ($\textsf{V}=5,000$).
They correspond to the most frequent words in the 1,010 privacy policies dataset.
Among them, 602 words cannot find pre-trained \textsf{word2vec} embeddings, we thus randomly initiate the embeddings using a standard normal distribution.

Next we feed the sentence one word at a time to a bi-directional recurrent neural network (forward layer colored in blue, backward layer colored in green, see Figure~\ref{fig:model}).
A recurrent neural network (RNN) corresponds to a language model, where the goal is to predict the next word given its previous words. The probability of the entire sequence $p(\bm{x})$ is represented in Eq.(1), whereas the individual probability $p(x_t|x_1,\cdots,x_{t-1})$ is calculated by RNN.  
\begin{align*}
p(\bm{x}) = \prod_{t=1}^{\textsf{N}} p(x_t|x_1,\cdots,x_{t-1}) \numberthis
\end{align*}

A recurrent neural network operates on a sequence of words and creates a hidden state representation $\bm{h}_t \in \mathbb{R}^d$ for the word at time step $t$.
It learns a function of the form $\bm{h}_t=f(\bm{h}_{t-1}, \bm{x}_t)$, where $\bm{h}_{t-1}$ is the hidden state representation of the previous time step and $\bm{x}_t$ is the input word embedding of the current time step.
Both the Long Short-Term Memory (LSTM) networks and Gated Recurrent Unit (GRU) networks are variants of the recurrent neural networks. 
They correspond to different gating mechanisms, hence different $f(\cdot)$.
This work specifically focuses on using GRU to produce the hidden state representations, where  $\bm{h}_t=\text{GRU}(\bm{h}_{t-1}, \bm{x}_t)$.
GRUs have seen considerable success in recent NLP applications~\cite{Luong:2016}.
It uses two neural gates to control the flow of information, where $\bm{i}_t \in \mathbb{R}^d$ and $\bm{r}_t \in \mathbb{R}^d$ respectively represent the \textit{input} and \textit{reset} gate.
$\bm{c}_t \in \mathbb{R}^d$ is sometimes referred to as the \textit{cell} value and $\bm{h}_t \in \mathbb{R}^d$ is the \textit{hidden} state representation we are interested in.
\begin{align*}
\bm{i}_t &= \sigma(\bm{W}^i\bm{x}_t + \bm{U}^i\bm{h}_{t-1} + \bm{b}^i) \numberthis\\
\bm{r}_t &= \sigma(\bm{W}^r\bm{x}_t + \bm{U}^r\bm{h}_{t-1} + \bm{b}^r) \numberthis\\
\bm{c}_t &= \tanh(\bm{W}^c\bm{x}_t + \bm{U}^c\bm{h}_{t-1} + \bm{b}^c) \numberthis\\
\bm{h}_t &= \bm{i}_t \odot \bm{c}_t + (1-\bm{i}_t) \odot \bm{h}_{t-1} \numberthis
\end{align*}

In the above equations, $\bm{W}^i, \bm{W}^r, \bm{W}^c$ and $\bm{U}^i, \bm{U}^r, \bm{U}^c$ are parameters, $\bm{b}^i, \bm{b}^r, \bm{b}^c$ are biases;
$\odot$ corresponds to the element-wise product of two vectors;
$\sigma(\cdot)$ is the sigmoid function; $\tanh(\cdot)$ is the hyperbolic tangent function. 
They are applied element-wise to the vectors.
We experiment with a bi-directional neural network, where in the forward-pass, $\text{GRU}_1$ admits words from the sentence beginning to end (Eq.(6)), and in the backward-pass, $\text{GRU}_2$ admits the word sequence reversely (Eq.(7)).
The generated hidden states are colored in blue (forward pass) and green (backward pass) respectively in Figure~\ref{fig:model}.
\begin{align*}
\overrightarrow{\bm{h}}_t=\text{GRU}_1(\overrightarrow{\bm{h}}_{t-1}, \bm{x}_t) \numberthis\\
\overleftarrow{\bm{h}}_t=\text{GRU}_2(\overleftarrow{\bm{h}}_{t-1}, \bm{x}_t) \numberthis
\end{align*}

The hidden state generated in the forward pass ($\overrightarrow{\bm{h}_t}$) is expected to carry over semantic information from beginning of the sentence to the current time step; similarly $\overleftarrow{\bm{h}_t}$ encodes information from the current time step to end of sentence.
We concatenate the two vectors of each time step $[\overrightarrow{\bm{h}_t}, \overleftarrow{\bm{h}_t}]$ and feed it to a densely connected layer to create a unified representation $\bm{g}_t \in \mathbb{R}^l$ for each word (colored in red in Figure~\ref{fig:model}. 
\begin{align*}
\bm{g}_t = \tanh (\bm{W}[\overrightarrow{\bm{h}_t}, \overleftarrow{\bm{h}_t}] + \bm{b}) \numberthis
\end{align*}

\noindent where $\bm{W}$ and $\bm{b}$ are parameters.
Using the vector representation $\bm{g}_t$, we learn to complete two tasks: 
first, $\bm{g}_t$ is used to predict the next word using a softmax activation function (Eq.(9)), where $p(y_t=j|\bm{g}_t)$ is the probability that the next word $y_t$ is predicted as the $j$-th word in the vocabulary; 
second, $\bm{g}_t$ is employed to predict if the current word is vague or not, where $p(c_t=k|\bm{g}_t)$ is the probability of the current word being vague ($k=1$) or not ($k=2$).
\begin{align*}
p(y_t=j|\bm{g}_t) = \frac{\exp(\bm{w}_j\bm{g}_t)}{\sum_{j'=1}^{\textsf{V}} \exp(\bm{w}_{j'}\bm{g}_t)}  \numberthis\\
p(c_t=k|\bm{g}_t) = \frac{\exp(\bm{w}_k\bm{g}_t)}{\sum_{k'=1}^{\textsf{C}} \exp(\bm{w}_{k'}\bm{g}_t)}  \numberthis
\end{align*}

\begin{figure}[t]
\centering
\includegraphics[width=3.3in]{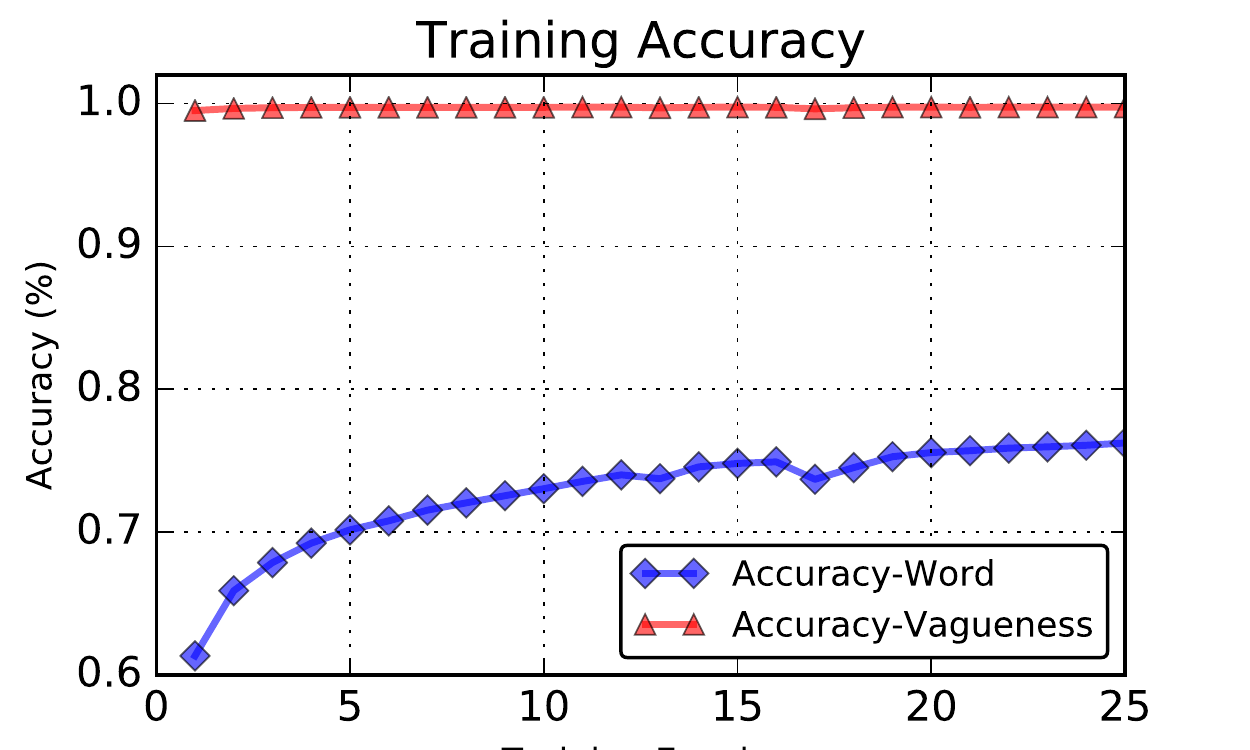}
\caption{Training accuracy across 25 epochs.}
\label{fig:train_accuracy}
\vspace{-0.1in}
\end{figure}

We use $\theta$ to represent all the trainable parameters in the aforementioned deep neural network. 
The above model can be trained in an end-to-end fashion using stochastic gradient descent. In particular RMSProp is used for parameter estimation, which has been shown to perform well in sequence learning tasks.
During training, the model parameters are iteratively updated so as to minimize the negative log likelihood of the training data $\mathcal{L}(\theta)$.
\begin{align*}
\mathcal{L}(\theta) =& - \alpha \sum_{i=1}^{\mathsf{S}} \sum_{t=1}^{\mathsf{N}} \sum_{j=1}^{\mathsf{V}} \log p(y_t=j|\bm{g}_t; \theta)\\
& - \beta \sum_{i=1}^{\mathsf{S}} \sum_{t=1}^{\mathsf{N}} \sum_{k=1}^{\mathsf{C}} \log p(c_t=k|\bm{g}_t; \theta) \numberthis
\end{align*}

\noindent where $\mathsf{S}$=107,076 is the total number of sentences in our dataset, $\mathsf{N}$=50 is set to be the maximum number of words per sentence, $\mathsf{V}$=5,000 is the vocabulary size, $\mathsf{C}$=2 is the number of categories (i.e., vague or not).
$\alpha$ and $\beta$ are scalar coefficients used to indicate the weights of the components in the leanring objective.
They are empirically set to $\alpha=1$ and $\beta=2$ in our study.
This means that the system is subject to heavier penalty when it mispredicts the word vagueness.
We set the dimensionality $d=512$ and $l=200$.
The deep neural network finally produces a 200-dimension vector representation for each word in the dataset.
The model is trained for 25 epochs.
Accuracy of predicting the identity of the next word (``Accuracy-Word'') and accuracy for predicting the word vagueness (``Accuracy-Vagueness'') are plotted in Figure~\ref{fig:train_accuracy}. 
The ``Accuracy-Vagueness'' curve saturates after the first couple of epochs, suggesting word-level binary prediction is not a difficult task, whereas the ``Accuracy-Word'' curve increases steadily across all the training epochs.

%Note that our goal of this paper is not on building an automatic system to predict if an individual word is vague or not. 
%Instead, our target of this work is to use the (incomplete) list of vagueness terms as a partial guidance to build a representation for each word in the privacy policies.
%The representation is in the form a dense vector $\bm{h}_t \in \mathbb{R}^d$, automatically learned using the deep neural networks.
%Each dimension in the vector works like an indicator of certain semantic information encoded in the text.
%We leverage on a recently developed visualization tool to illustrate what semantic information has been encoded in this vector representation.

\begin{figure*}
\centering
\includegraphics[width=7in]{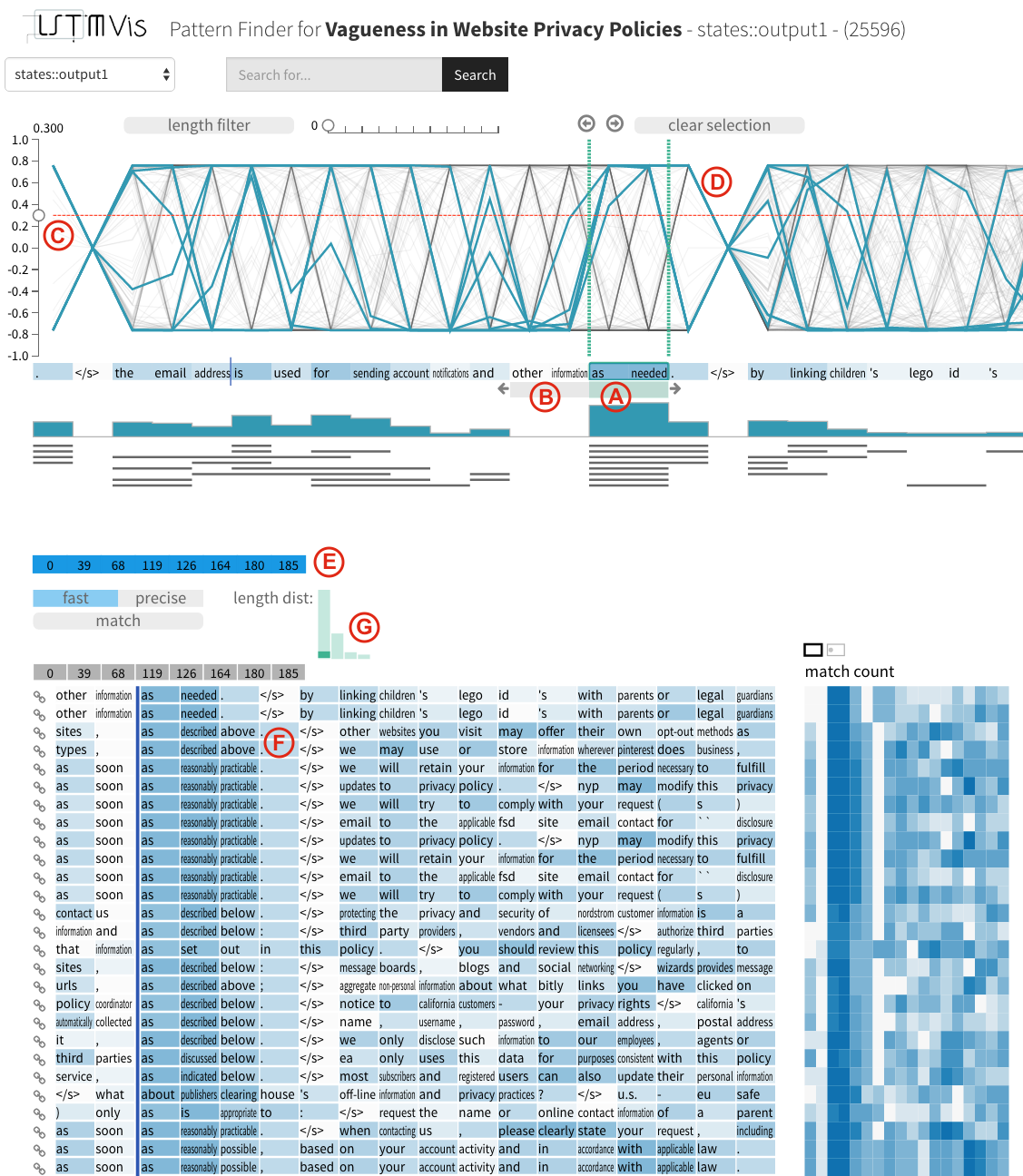}
\caption{Visualization of the vector representations using \textsf{LSTMVis}.}
\label{fig:vis}
\vspace{-0.1in}
\end{figure*}

\section{Visualization}
\label{sec:vis}

\begin{table*}
\setlength{\tabcolsep}{7pt}
\renewcommand{\arraystretch}{1.2}
\centering
\begin{tabular}{|l|l|l|}
\hline
\textbf{under the same circumstances} & \textbf{necessary or appropriate to} & \textbf{personally-identifying information}\\
\hline
under the following conditions & necessary to & personal information\\
under the following circumstances & required to & access information\\
under the circumstances & otherwise permitted by & financial information\\
in any case & your right to & aggregate information\\
in this case & & contact information\\
\hline
\end{tabular}
\caption{Example similar phrases identified by the visual tool. The given phrases are shown in bold. Note that the similar phrases are hand-picked. Not all system identified phrases are closely related to the given phrases.}
\label{tab:phrases}
\vspace{-0.1in}
\end{table*}

The deep neural network presented in the previous section creates a 200-dimension vector representation for each word in the privacy policy dataset.
The vectors are colored in red in Figure~\ref{fig:model}.
These vector representations resemble the feature vectors we normally obtain through a linear model (e.g., SVM or maximum entropy) or dimensionality reduction approach (e.g., SVD).
They could be used in downstream classification tasks such as predicting if a piece of text is vague or not. 
However, because of the lack of such large-scale datasets and consistent annotation guidelines for vagueness prediction of privacy policies, we choose not to perform empirical evaluation on the datasets.
Instead, we seek to interpret the learnt vector representations and explore what information is encoded in the 200-dimension vectors.

Researchers strive to understand the neural models in natural language processing.
Very recently, Li et al.~\shortcite{Li:2016} develop strategies to understand the model compositionality. 
That is, how sentence meanings are built from the meanings of words and phrases.  
The approach measures the ``salience'' of each dimension based on how much it contributes to the final decision, which is approximated using first-order derivatives.
Strobelt et al.~\shortcite{Strobelt:2016} present a visual analysis tool named \textsf{LSTMVis}\footnote{lstm.seas.harvard.edu}. %TODO
The tool explores the hidden state dynamics of a recurrent neural network.
It allows the user to select an input phrase and find similar phrases in the dataset that demonstrate similar hidden state patterns.
We adopt \textsf{LSTMVis} in our study and import the vector representations produced in the previous section.
The visualization is presented in Figure~\ref{fig:vis}.

The interface consists of two views: the \textit{select} view corresponds to the upper panel ((A) to (D)) and the \textit{match} view corresponds to the lower panel ((E) to (G)).
All sentences in the dataset are concatenated into a meta word sequence and delimited by the special symbol $\langle/\mathsf{s}\rangle$.
Each word is represented using a fixed-width box; if words do not fit into the box, they are squeezed. 
Users are provided with buttons to move forwards or backwards with the word sequence, as well as a search box (disabled for now) to directly jump to certain text region.
Each vector dimension corresponds to a line in the \textit{select} view.
Because our vector representation contains 200 dimensions, there are 200 lines in the figure, numbered from 0 to 199.

The user starts by selecting a phrase in the word sequence (e.g., ``as needed,'' see (A)).
This action turns on a set of vector dimensions (represented as $S_1$), where ``turn on'' means the cell value of the dimension, in both of the selected word positions, is greater than a threshold (default to 0.3, see (C)).
The gray slider (see (B)) further allows the user to select a few context words (e.g., ``other information'') that surround the current selected phrase.
Similarly, this action turns on a second set of vector dimensions (represented as $S_2$).
Note that our goal is to identify the dimensions that uniquely characterize the selected phrase (``as needed'') but not the surrounding words.
As a result, the intersection of the two sets of dimensions $|S_1 \cap S_2|$ are the ones we wish to focus on.
These dimensions are listed in the interface (see (E)).

In the \textit{match} view, the visual tool continues to search for text regions where the same set of vector dimensions ($|S_1 \cap S_2|$) have been turned on. 
The text regions are further ranked by the inverse of number of additional ``on'' cells $|S_1 \cup S_2|$ and the length of the text region.
The top phrases are listed on the interface (see (F)) with length distribution plotted (see (G)).
The color intensity is used to signal the value of $|S_1 \cap S_2|$.
For the selected phrase (``as needed''), we observe that several syntactically and semantically similar phrases have been selected, including ``as is appropriate to,'' ``as described below,'' ``as reasonably possible,'' ``as reasonably practicable,'' and ``as set out in.''
Several similar examples are presented in Table~\ref{tab:phrases}.
These findings suggest that even in the relatively restricted domain of website privacy policies, a large number of text variations exist.
They use different text expressions to represent the same or similar meanings.
It is thus left to be seen if creating a comprehensive list of vague terms is feasible given the richness and complexity of natural language.

\section{Conclusion}
\label{sec:conclusion}

In this work we attempt to computationally model the vagueness of privacy policies using deep neural networks.
The neural networks learn to generate vector representations for words in the privacy policies.
We explore visualization of the learnt vector representations, identify dimensions that could capture language specific characteristics, and present example phrases that potentially signal vagueness.
Our learned model and visualization allow researchers to explore the vagueness of natural language and perform knowledge discovery.
We expect future work will include empirical evaluations on vagueness datasets and use the vagueness prediction results to assist legal counsels to clarify the privacy text, as well as raise public awareness of the vague terms as presented in the website privacy policies.

\bibliographystyle{aaai}
\bibliography{vague}

\end{document}